\documentclass[10pt,journal,compsoc]{IEEEtran}

\ifCLASSOPTIONcompsoc
  \usepackage[nocompress]{cite}
\else
  \usepackage{cite}
\fi

\usepackage{amsmath}
\usepackage{amssymb}
\usepackage[pdftex]{graphicx}
\usepackage{color}

\newcommand{\norm}[1]{\left\Vert#1\right\Vert}
\newcommand{\abs}[1]{\left\vert#1\right\vert}
\newcommand{\Real}{\mathbb R}
\newcommand{\Integer}{\mathbb Z}

\newcommand\CC[1]{\mathbb{C}^{#1}}

\newtheorem{thm}{Theorem}
\newtheorem{cor}[thm]{Corollary}
\newtheorem{lem}[thm]{Lemma}
\newtheorem{prop}[thm]{Proposition}
\newtheorem{conj}[thm]{Conjecture}

\DeclareMathOperator{\diag}{diag}

\begin{document}

\title{A function space analysis of finite neural networks with insights from sampling theory }


\author{Raja Giryes,~\IEEEmembership{Senior Member,~IEEE}
\IEEEcompsocitemizethanks{\IEEEcompsocthanksitem R. Giryes is with the School
of Electrical Engineering, Tel Aviv University.\protect\\
E-mail: raja@tauex.tau.ac.il}
\thanks{Manuscript received May 19, 2020; revised September 2, 2021; accepted February 9, 2022.}}

\markboth{To appear in Transactions on Pattern Analysis and Machine Intelligence,~2022}%
{R. Giryes: A function space analysis of finite neural networks}

\IEEEtitleabstractindextext{%
\begin{abstract}
This work suggests using sampling theory to analyze the function space represented by interpolating mappings. While the analysis in this paper is general, we focus it on neural networks with bounded weights that are known for their ability to interpolate (fit) the training data. First, we show, under the assumption of a finite input domain, which is the common case in training neural networks, that the function space generated by multi-layer networks with bounded weights, and non-expansive activation functions are smooth. This extends over previous works that show results for the case of infinite width ReLU networks. Then, under the assumption that the input is band-limited, we provide novel error bounds for univariate neural networks. We  analyze both deterministic uniform and random sampling showing the advantage of the former. 
\end{abstract}

}

\maketitle

\section{Introduction}

Recently, it has been shown that neural networks with a univariate output and bounded weights perform a smooth interpolation between their training data \cite{Savarese19How,Williams19Gradient,Ongie2020Function}. These works provide an extension to many recent results that have studied the approximation power of neural networks. While in the general universal approximation theory, either in the infinite width case \cite{Cybenko89Approximation, Hornik98Multilayer} or the finite width case \cite{Lu17Expressive}, it is shown that virtually any function may be approximated, the new results demonstrate that by adding constraints on the network weights, we get a smaller function space although the width of the network is infinite.  

An interesting question that may arise as a follow up to these works that focused on the \textit{approximation power} of the network is whether we may use their results to get new \textit{estimation error} bounds for networks trained on $n$ data samples. One intriguing phenomenon of neural networks is that for ``natural good data'' they both overfit the training data and generalize well at the same time, while for random ``bad'' data they just perform memorization with no generalization \cite{Zhang17Understanding}. This phenomenon hints that the generalization of the network depends also on the structure of the input data and not only on the network parameters.  

In this work, we focus on the case of data that is generated by band-limited functions and that the neural network reaches a zero training error, i.e., interpolating the data. This behavior is observed in practice and proved in various recent works \cite{jacot2018neural,chizat2018note,mei2018mean,lee2019wide,arora2019exact}.
We show for a neural network that when it interpolates (overfits) the training data, its squared error scales as $O(1/n^{(d+2)/d})$, where $n$ is the size of the training data and $d$ is the dimension of the input. Note that our result suggests that for large $d$ the squared error scales as $1/n$, and therefore the error without the square scales as $1/\sqrt{n}$, which coincides with the known bounds for the error of neural networks. Yet, for low-dimensional inputs, e.g., $d=1$, our bounds are much better than the existing ones. This is possible by our assumption on the input data and by incorporating their dimension in the analysis.
Moreover, our result provides a concrete example where the memorization of the network helps its generalization. Note that it naturally excludes the case of random data, which have an infinite bandwidth. 
As we shall see hereafter in the proof of our results, the fact that the network fits all the training examples is a key element in its ability to get low error for all the other points of the function that generated the data. 

The contribution of our work is twofold. First, we show that the function represented by a finite width network with bounded weights have a bounded total variation of its first derivative, i.e., $\int_{-\pi}^\pi f''(x)dx < \infty$, where $[-\pi,\pi]$ is assumed to be the input domain. This shows that finite networks perform a smooth interpolation of their training data. This extends over previous works that have been limited to infinite width networks. 
The second is providing generalization results both for infinite width networks and finite width ones. We use tools from sampling theory to analyze the error of the network both in the case of deterministic uniform sampling (Theorem~\ref{thm:uniform_sample_rec}) and the more realistic case of random sampling (Theorem~\ref{thm:nonuniform_sample_rec}). Then in Theorem~\ref{thm:uniform_rhombus_rec} we extend the results to multivariate functions (under the assumption of uniform sampling).  Notice that the analysis performed in these theorems is general to any mapping that interpolates the training data. Yet, we put the focus on neural networks in this paper because of the following two important properties: (i) they are known to be able to interpolate the data; and (ii) when the weights of the network are bounded then the frequencies of the mapping represented by the network decay rapidly (as we prove in the first part of the paper), which is one of the characteristics that the mapping should satisfy for our analysis to hold.

\section{Related work}

A relationship between network representation and a given function space was shown in \cite{Candes99Harmonic, Sonoda17Neural}. In particular, these works focused on the ridgelet transform. The first studied the approximation power of networks with some special activation function using ridgelets. The second presented a connection between neural networks with ReLU activation and the ridgelet transform. They demonstrated that such networks satisfy the universal approximation property.
Another line of works showed that networks learn first lower frequencies in the data \cite{Rahaman19Spectral, Xu19Training,Bar2022Spectral}. 
Another paper \cite{Poggio2020Theoretical} analyzes the impact of gradient descent on the network approximation power.
The work in \cite{li2021why} studies the gap between the sample complexity required for training a fully connected and a CNN. They show that CNN may require significantly less samples compared to a fully connected network. This is different than this paper that focuses on the impact of the network smoothness on its generalization performance.

The works in \cite{Savarese19How,Williams19Gradient,Ongie2020Function} have shown that shallow infinite width networks with bounded weights perform a smooth (spline) interpolation of the training data. 
Another connection between neural networks and splines was exhibited in \cite{Balestriero18Spline}. It focused on the specific case of max affine splines and used them to show a relationship between template matching and networks.

A connection between adding a regularization on the weights of the network and their generalization was shown in various works. While classic generalization error bounds for neural networks presented a dependency on the number of parameters in the network \cite{Barron1994Approximation}, Rademacher complexity (RC) based analysis showed that by bounding the norm of the weights, the generalization error is independent of the network width \cite{Neyshabur15Norm,Golowich18Size}. The work in \cite{Zhou18Understanding} provided improved generalization bounds, which depend on the log of the product of the network weights instead of only the products. Yet, the deficiency of these bounds is their independence of the input data; thus, they do not capture cases such as overfitting of random data \cite{Zhang17Understanding}. 

Margin based approaches, which take into account also the input distribution, mitigate this issue \cite{Sokolic17Robust,Bartlett17Spectrally,li2019tighter}. Note that ``$\ell_2$ regularization does not significantly impact margins or generalization'' \cite{Bartlett17Spectrally}, where the analysis here depends on the consequence of this regularization. 
Thus, these approaches are complementary to our analysis. 
Bounding the weights is also shown useful under the kernel (RKHS) assumption \cite{Wei19Regularization}, which is not required in our work. Generalization error bounds for data that is separable under some random feature network or kernel is shown \cite{Cao2020Generalization,Cao19Generalization}. This is different than our work, which assumes a more realistic assumption on the mapping function that is generating the data, namely, that it is band-limited (i.e., smooth).

The contribution of this work is also relevant to general sampling theory. Indeed, many results have been developed in this field for the reconstruction performance of interpolation techniques from both uniform and non-uniform samples \cite{Jerri77Shannon,Unser97Generalized,Margolis08Nonuniform}. Yet, all these results assume that the interpolating function belongs to the space of the target functions (e.g., only generating band-limited functions in  the reconstruction). In our case, we do not have this assumption as the neural network does not necessarily generate band-limited functions. Thus, we develop theoretical reconstruction guarantees for this setting.

\section{Neural networks and sampling theory preliminaries}

This section surveys some preliminaries of neural networks and sampling theory. Readers that are familiar with these topics may skip to the next section. 

Any neural network training relies on a given input dataset $\{(x_i, y_i)\}_{i=0}^{n-1}$ with $n$ pairs of data sample $x_i$ and label $y_i$.  
In general, the input space of a neural network is limited, i.e., $x$ is sampled just from a specific interval of interest (for example, in images the pixel values are only in the range $[0,255]$). Without loss of generality, we will assume for the simplicity of the presentation that $x \in [-\pi, \pi]$. 
In this case, we can arbitrarily define the values of $f(x)$ outside this interval (we do not sample the function there and therefore it does not affect the data generation and the network trained).
We specifically select a periodic continuation of $f$ such that $f(x) = f(x+2\pi)$.

Since we assume that $f$ is bandlimited, then $f$ must be also smooth and thus this assumption implies that $f(-\pi) = f(\pi)$. Notice that this assumption does not limit us in any way as if this is not the case, there are various ways to mitigate this issue. For example,  in the case that $f(x) - f(x +2\pi)$ is not too large, we may extend the function a bit beyond $x=2\pi$ in a smooth way such that it will remain band limited and satisfy the periodicity assumption.  
Another popular alternative is using a symmetric expansion of $f$ (copying a mirrored version of $f$ in the interval $[-\pi,\pi]$ to the interval $[\pi,3\pi]$, which enforces having $f(-\pi) = f(3\pi)$ due to the mirroring) before applying the periodic extension. This just changes the integral limits when calculating the Fourier coefficients of $f$ and requires replacing the DFT (which we use hereafter) with DCT (Discrete Cosine Transform) \cite{Grishin04Fast}. 

Since $f$ is periodic, we may calculate its Fourier coefficients 
\begin{eqnarray}
\label{eq:Fourier_coef_calc}
c_k = \frac{1}{(2\pi)^d}\int_{\norm{x}_\infty \le \pi} f(x)e^{-j x^T k}dx,
\end{eqnarray}
where $k \in \Integer^d$. If $f$ is bandlimited (also known as trigonometric polynomial \cite{Rudin87Real}) then $c_k =0$ if $\exists i$ such that $k[i] >K$. Thus,
\begin{eqnarray}
f(x) = \sum_{-K\le k[i] \le K } c_k e^{j x^T k}.
\end{eqnarray}
Note that we sum over all the combinations in which $\abs{k[i]} \le K$.

Using the sampling theorem for bandlimited periodic signals, we may recover $f(x)$ using just $n \ge N \triangleq (2K+1)^d$ samples $\{f(x_i)\}_{i=0}^{n-1}$. For completeness, and as it will help us later in the derivations, we briefly describe here this result.

{\bf Uniform sampling.} We start with reconstruction using uniform sampling. Assume that our sample points are on the grid $$[\frac{2\pi i_1}{2K+1},\frac{2\pi i_2}{2K+1},\dots,\frac{2\pi i_d}{2K+1}],$$ where $i_l =0, \dots, 2K$ for $l=1, \dots, d$. 
In the one dimensional case ($d=1$), we have \begin{eqnarray}
\label{eq:1d_uniform_samples}
f(x_i) = \sum_{k =-K}^K c_k e^{\frac{j2 \pi k i}{2K+1} }.
\end{eqnarray}
Denoting by $c$ the vector that contains the Fourier coefficients in it and by $y$ the vector that contains the values of $f(x_i)$, we may rewrite \eqref{eq:1d_uniform_samples} as (see \cite{Margolis08Nonuniform})
\begin{eqnarray}
\label{eq:y_Fc}
y = F^*c,
\end{eqnarray}
where $F\in \CC{N \times N}$ is the DFT (Discrete Fourier Transform) matrix, whose columns (in 1D) are of the form $\{e^{j2\pi k\frac{i}{2K+1}}\}_{k=-K}^K$,
and $F^*$ is its conjugate transpose, which is also its inverse (up to a scale factor $1/N$) because the rows of $F$ are orthogonal to each other. Notice that the same holds true for the multi-dimensional case ($d>1$) and then $F$ is simply the $d$-dimensional DFT (in this case, we can also cast $c$ in a vector representation). Having this relationship, we can recover the vector $c$, and thus the whole function $f$, from $y$ by computing $c = \frac{1}{N}F y$.

{\bf Oversampling.} Notice that if the number of measurements that we have are $n > N$, then we still have the relationship in \eqref{eq:y_Fc} but in this case $F \in \CC{N \times n}$ is a DFT (tight) frame, whose columns are of the form $\{e^{j2\pi k\frac{i}{n}}\}_{k=-K}^K$ (in 1d).
Since the rows of $F$ are orthogonal in this case as well (also for $d>1$), we still have that $\frac{1}{n}FF^* = I$ and thus we can reconstruct the function $f$ using $c = \frac{1}{n}Fy$ as before. 

Notice that due to the redundancy that we have in the measurements, we may use other DFT operators to reconstruct $c$. In particular, for any $\tilde{N}$ and $n\ge \tilde{N} \ge (2K+1)^d$, we can simply pad $c$ with zeros, which yields the relationship $y = F^*c$ for the DFT frame $F\in \CC{\tilde{N} \times n}$ (which is the standard DFT transform if $n = \tilde{N}$). As before, we can reconstruct the Fourier coefficients by $c = \frac{1}{n}Fy$. We abuse notation here and elsewhere  denoting by $c$ also the padded representation. The use will be clear from the context.

{\bf Non-uniform sampling.} In many cases, we get just a random (non-uniform) set of samples of the space. In this case, the set of input points $\{x_i \}_{i=0}^{n-1}$ do not lie on the grid but are randomly spread in $[-\pi, \pi]^d$. The sampled points obey
\begin{eqnarray}
\label{eq:1d_nonuniform_samples}
f(x_i) = \sum_{k =-K}^{K} c_k e^{ j k^T x_i }.
\end{eqnarray}
Writing \eqref{eq:1d_nonuniform_samples} in a matrix form yields $f=Dc$, where the rows of $D$ are $\{e^{jkx_i}\}_{k=-K}^{K}$ (in the  1D case). Notice that $D \in \CC{n \times N}$ is very similar to the DFT inverse transform ($F^*$) but with the difference that its rows correspond to random frequencies unlike $F^*$ whose rows have equi-spaced frequencies (that leads to the orthogonality property). Notice that also here we may pad $c$ with zeros and thus have $D \in \CC{n \times \tilde{N}}$ in a similar way to the oversampling case. If $\tilde{N} = 2\tilde{K}+1$ for some $\tilde{K} \ge K$ then the rows of $D$ are $\{e^{jkx_i}\}_{k=-\tilde{K}}^{\tilde{K}}$ (in the 1D case).

If $D$ has a full column rank  (which is the case of many random sampling schemes \cite{Jerri77Shannon,Unser97Generalized,Margolis08Nonuniform}), i.e., invertible, then we may again reconstruct the function $f$ by computing $c = D^{\dag}y$, where $D^{\dag} = (D^* D)^{-1}D^*$ is the pseudo-inverse of $D$. Although we can get perfect reconstruction also with random sampling, its disadvantage is the noisy case, where we get noise amplification that depends on the ratio $\kappa$ between the largest and smallest non-zero singular value of $D$ (the condition number of $D^* D$). This ratio is dependent also on the ratio between $n$ and $\tilde{N}$ \cite{Farrell11Limiting,Haikin17Random}. We use it hereafter  to provide error bounds for neural networks with randomly sampled training data.

\begin{figure}
\begin{center}
   \includegraphics[width=0.48\textwidth]{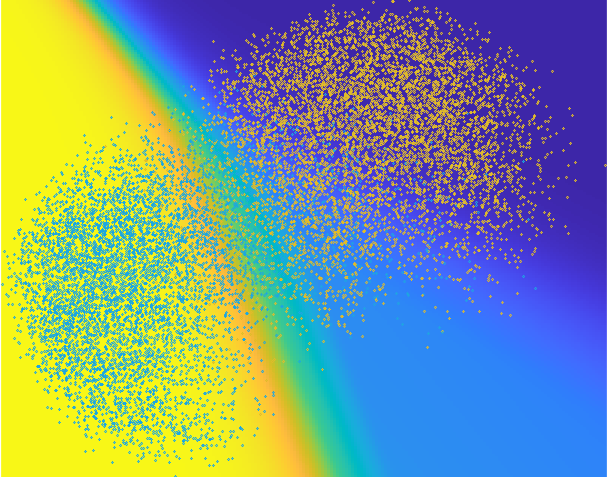}
   \includegraphics[width=0.48\textwidth]{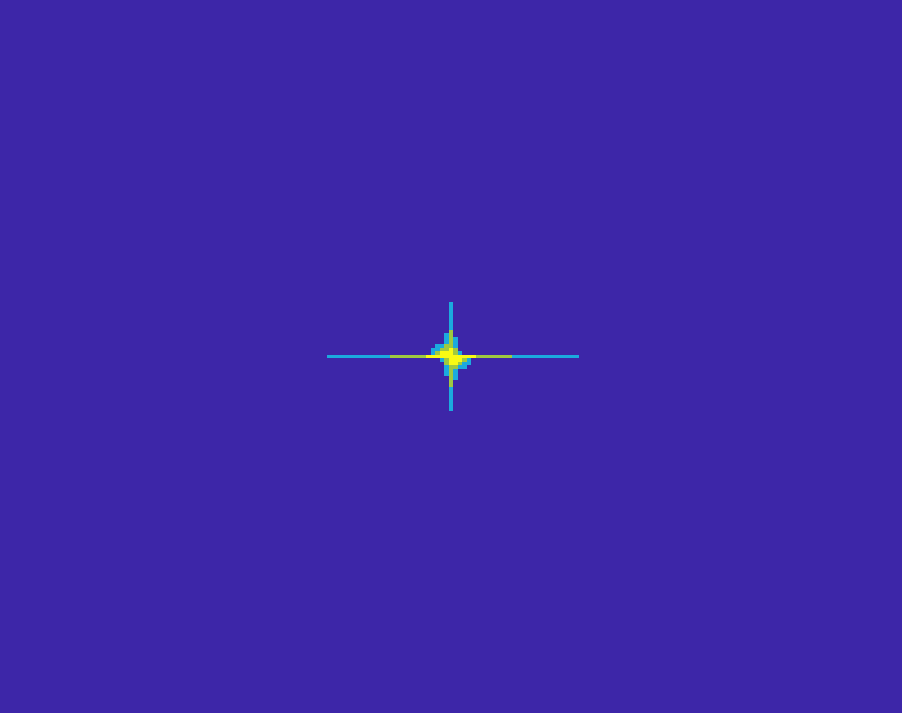}
\end{center}
   \caption{Top: The mapping function learned by the network for the two-dimensional projections of the digits 1 and 7 (the blue and yellow scattered points). The yellow and blue colors in the mapping represent outputs for 1 and 7 respectively. Bottom: The Fourier transform of the above mapping. We present in bright yellow the frequencies corresponding to 95\% of the energy of the mapped function. Dark yellow and bright blue corresponds to the frequencies the complement to 98\% and 99\% of the energy. It can be clearly seen that the mapping is (approximately) band limited. }
\label{fig:mnist2_mapping}
\end{figure}

{\bf Smoothness of neural network mappings.} 
The underlying assumption in this work is that the mapping function of the (true) data $f$ is band limited (or having rapidly decaying high frequencies). To justify this assumption, we train a neural network on data of two digits ($1$ and $7$) from MNIST 
and show that the learned mapping that overfits (most of) the data has fast decaying frequencies. To be able to make the visualization of the network mapping, we project the images of the digits to a two dimensional subspace using PCA (using the two largest components). We then train a simple network $2 \rightarrow 1000 \rightarrow 1000 \rightarrow 2$ with ReLU as the activation function and softmax at the end. The network reaches an accuracy of $\sim92\%$. Figure~\ref{fig:mnist2_mapping} shows the learned mapping and the coefficients of its Fourier transform that hold 95\% (bright yellow), 98\% (dark yellow), and 99\% (bright blue) of the energy of the mapping. To generate this plot, we calculate on a 2D grid the outputs of the network $f$ and then applied FFT (Fast Fourier Transform) to this grid. The plot shows the locations of the coefficients with the largest magnitudes.
It can be clearly seen that the mapping of this data is (approximately) band limited. 

Notice that the assumption that the mapping function of the true data should have fast decaying frequencies (i.e., the decision boundaries in it should be relatively smooth as is the case in Figure~\ref{fig:mnist2_mapping}) is a hidden assumption in other works, e.g. the recent work by Ghorbani et al. \cite{ghorbani2021linearized} that suggests that the decision boundary in linear networks is of lower order polynomial (which implies that the mapping they represent will have fast decaying high frequencies). 

While the analysis in the paper assumes for the simplicity of the analysis only the “pure” band limited mapping case, we explain at the end how it may be extended also to the case of approximately band limited mappings (i.e., with fast decaying high frequencies). 

\section{The function space of bounded finite neural networks}
\label{sec:net_func_space}

The work in \cite{Savarese19How} proved that any function $\phi$ represented by a two layer overparameterized (with number of parameters going to infinity) ReLU network  with univariate input and output has a bounded total variation in their first derivative as the bound on the network norm imposes a constraint on
\begin{equation}
\max \left( \int_x \phi''(x)dx , \phi'(\infty) + \phi'(-\infty) \right).
\end{equation}
They have shown that this implies a spline interpolation (of at least order one, i.e., linear) between the training data, which the network overfitted (which is possible due to its overparameterization). The work in \cite{Williams19Gradient} have extended their results showing that the network performs a second order (cubic) spline interpolation between the data points under some assumption on the initial weights and the optimization process. The result of \cite{Savarese19How} have been extended in \cite{Ongie2020Function} to the case of multi-dimensional input. They have shown that in this case, the functions represented by the network have a bounded $\mathcal{R}$-norm, which is related to the Radon transform of the represented function.

Notice that the existing works \cite{Savarese19How,Williams19Gradient,Ongie2020Function} assume shallow networks with infinite width. We show here that under the assumption that the input domain is bounded (as is the common case with neural networks training), then neural networks with bounded norm approximate functions that have a bounded derivative and thus also total variation in the second derivative. 
These papers show that the optimization is precisely controlling the $\ell_1$ norms (of the second derivatives in the case of dimension 1) in two-layer infinite-width networks, leading to a minimum-norm interpolating solution. We take a different approach that do not assume a specific algorithm for training the network except of that it leads to fitting the training data and having bounded weights. We use that to show that a similar quantity to the one studied in \cite{Savarese19How,Williams19Gradient,Ongie2020Function} is bounded, but not that it dictates the solution (as we do not assume a specific algorithm).

Denote by $\sigma_i$ the non-linearity in the network at the $i$th layer and by $W_i$ and $b_i$ the weights and biases there. Then, we may write a feed-forward network with $L$ layers as
\begin{eqnarray}
\nonumber
\phi(x) = \sigma_L(b_L + W_L \sigma_{L-1}(   \cdots \sigma_2(b_2 + W_2 \sigma_1(b_1 + W_1 x)).
\end{eqnarray}
If we denote by $z_i$ the output of the $i$th layer, then we can write the above recursively as
\begin{eqnarray}
z_i = \sigma_i (b_i + W_i z_{i-1}),
\end{eqnarray}
where $z_0 = x$ and $z_L = \phi(x)$. For such a network we prove the following proposition, which is an extension of the result in \cite{Sokolic17Robust}.

We rely on a result from \cite{Sokolic17Robust} that shows the relationship $\norm{\frac{d\phi}{dx}} \le \prod_i \norm{W_i}_F$. Yet, that work presents this result only for networks with ReLU, Sigmoid or hyperbolic tangent as non-linearity and without biases. The following proposition presents this result also for networks with biases and other non-expansive activation functions.

\begin{prop}
\label{lem:bounded_norm}
Let $\phi(x)$ be a feed-forward network with an input $x$, non-expansive non-linear function $\sigma_i$ and weights and biases $\{W_i\}_{i=1}^L$ and $\{b_i\}_{i=1}^L$. Then, we have
\begin{eqnarray}
\label{eq:Jacobian_W_prod_bound}
\norm{\frac{d\phi}{dx}} \le \prod_{i=1}^L \norm{W_i} \le \prod_{i=1}^L \norm{W_i}_F,
\end{eqnarray}
where $\norm{\cdot}$ and $\norm{\cdot}_F$ are the spectral and Frobinius norms respectively. Notice that the product $\prod_{i=1}^L \norm{W_i}_F^2$ can be upper bounded by the sum $\sum_{i=1}^L \norm{W_i}_F^2$.
\end{prop}

\emph{Proof.}
For calculating the Jacobian  $\frac{d\phi}{dx}$, we may use the chain rule (as used in back-propagation), getting
\begin{eqnarray}
\label{eq:Jacobian_chain_rule}
\frac{d\phi}{dx} = \frac{d\phi}{dz_{L-1}} \frac{dz_{L-1}}{dz_{L-2}} \cdots \frac{dz_{2}}{dz_{1}}
\frac{dz_{1}}{dx}.
\end{eqnarray}
Thus, using matrix norm inequalities we have
\begin{eqnarray}
\label{eq:Jacobian_chain_rule_norms}
\norm{\frac{d\phi}{dx}} = \norm{\prod_{i=1}^L \frac{dz_{i}}{dz_{i-1}}} \le \prod_{i=1}^L \norm{\frac{dz_{i}}{dz_{i-1}} }.
\end{eqnarray}
Now, notice that 
\begin{eqnarray}
\frac{dz_{i}}{dz_{i-1}} =   \diag(\sigma_i' (b_i + W_i z_{i-1}))W_i.
\end{eqnarray}
Since the spectral norm of the diagonal matrix $\diag(\sigma_i' (b_i + W_i z_{i-1}))$ is its maximal value and as this value is smaller or equal to $1$ (as we assume $\sigma$ is non-expansive), we have that 
\begin{eqnarray}
\norm{\frac{dz_{i}}{dz_{i-1}}} = \norm{\diag(\sigma_i' (b_i + W_i z_{i-1}))W_i} \le  \norm{W_i}.
\end{eqnarray}
Plugging this inequality in \eqref{eq:Jacobian_chain_rule_norms} and then using the known relationship between the spectral and the Frobenius norms, we get the desired result.
 \hfill $\Box$

To get a bound on the total variation of the second derivative we make the following simple observation: The discontinuities in the function approximated by the network are only due to the non-linear function in the network. Since the first derivative is bounded the ``jumps'' that occur in it are finite.  
Since we are dealing with a finite domain and a finite network, the number of such discontinuities is finite and therefore the integral over the second derivative is also finite (also known as the total variation of the first derivative).  

Notice that in the case of infinite network and infinite domain, we cannot make the above assumptions and therefore a more sophisticated approach as the one in \cite{Savarese19How} is required to give a bound on the total variation of the first derivative. Yet, their work does not apply to the finite network case as does our result here. Notice that for shallow networks, which is the case studied in \cite{Savarese19How,Williams19Gradient,Ongie2020Function}, the number of discontinuities in the network grows linearly with the width. In the deeper case, it grows faster (see analysis in \cite{Balestriero18Spline,Montufar14Number}) but is still bounded.

This provides us with the following corollary for finite neural networks with a univariate output.
\begin{cor}
\label{cor:bounded_lap}
Let $\phi$ be a finite multi-layer neural network with bounded weights (i.e., $\prod_{i=1}^L \norm{W_i}$ or $\prod_{i=1}^L \norm{W_i}_F$ are bounded) and non-expansive non-linearities that have a finite amount of discontinuities in their first derivative. Assume the training data is in the interval $[-\pi, \pi]^d$. Then the 
total variation of the derivative of this function,   $\int_{x \in [-\pi, \pi]^d} \Delta \phi(x)dx$, is finite, where $\Delta \phi(x) = \nabla^2 \phi(x)$ is the Laplacian of $\phi(x)$.
\end{cor}
\emph{Proof.}
Using Proposition~\ref{lem:bounded_norm} we have that all the partial derivatives of $\phi(x)$ are bounded in the domain $[-\pi, \pi]^d$. Since the network is finite and the discontinuities in the network derivative emerges from the non-linearities that have a finite amount of discontinuities in their first derivative, we have a finite amount of ``jumps'' in the interval $[-\pi,\pi]$ and all of them. The integral over the second derivative can be bounded by the difference between the largest and smallest first derivative of $\phi$ times the interval size plus the sum of the sizes of the jumps (as each is a delta function in the second derivative). As the first derivative and the amount of ``jumps''  are bounded, we have that $\int_{x \in [-\pi, \pi]^d} \nabla^2 \phi(x) dx$ is finite.
 \hfill $\Box$ 



\section{Sampling theory based error bounds}

We turn now to use the above findings to prove that a neural network with bounded norms can recover band-limited functions with very high precision both with uniform and non-uniform sampling, where the latter is the more common case when getting a training data for a neural network. 
The underlying assumption in the analysis here is that the labels $y_i$ are generated by a band limited function $f(x)$. We also assume that the neural network used interpolates the data and has bounded weights. Specifically, that $\prod_i \norm{W_i}_F$ is bounded (as we rely on Corollary~\ref{cor:bounded_lap} in our analysis). 

\subsection{A periodic representation of the neural network function}

Denote by $\tilde{\phi}_n : \Real \rightarrow \Real$ a function represented by a neural network that has bounded weights and is trained with $n$ training samples. 
While this function is defined for all $\Real$, for our data we are only interested in the output of the network in the domain $[-\pi, \pi]$.
Therefore, for analyzing the network estimation error compared to the function $f(x)$ in this domain, we can change $\tilde{\phi}_n$ arbitrarily as we wish outside of this domain. 

To be able to calculate a Fourier series, we define the function $\phi_n$, which is equal to $\tilde{\phi}_n$ in the domain $[-\pi, \pi]$ and is periodic outside of it with a period $2\pi$. Clearly, also in this case we may have that $\tilde{\phi}_n(x+2\pi) \ne \tilde{\phi}_n(x)$. Yet, as we discussed in the preliminaries section, this can be leveraged, for example, by using a symmetric extension and then the same analysis that we present below will remain the same but with a DCT replacing the DFT used in the analysis. Since both are orthogonal, the derived results remain the same. Thus, for simplicity we just assume a regular periodic extension.

Given that $\phi_n$ is periodic we may calculate its Fourier series 
\begin{eqnarray}
\phi_n(x) = \sum_{k\in \Integer} \zeta_k e^{j x k}, 
\end{eqnarray}
where $\zeta_k$ is calculated as in \eqref{eq:Fourier_coef_calc} (with $\phi_n$ instead of $f$).

\begin{figure}
\begin{center}
   \includegraphics[width=0.48\textwidth]{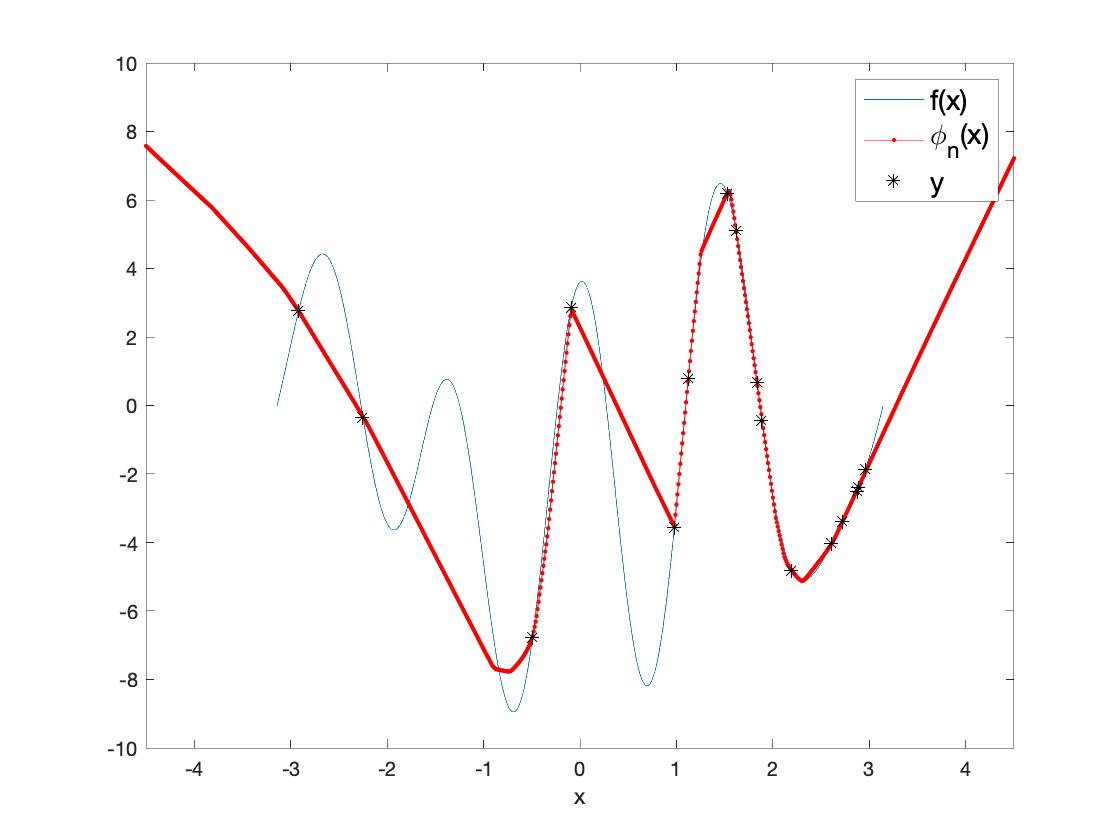}
\end{center}
   \caption{Approximation of a band-limited function $f(x)$ using a neural network $\phi_n$ trained using only $16$ training examples ($y$).}
\label{fig:reconstruction}
\end{figure}

Now, assume that the network has overfitted the data, i.e., $f(x_i) = \phi_n(x_i)$, then if $\phi_n$ is band-limited as $f$, then we get from sampling theory that $f = \phi_n$. 
In the case of uniform sampling, if the network function $\phi_n$ was exactly a spline, we could have used the result in \cite{Jacob02Sampling} to calculate the network error as a function of $n$. 
Yet, $\phi_n$ is not guaranteed to be band-limited and
as shown in \cite{Savarese19How,Williams19Gradient}, the connection between the points may be beyond ''linear''. Figure~\ref{fig:reconstruction} provides an example of a trained network output, where we get different types of interpolations between the training points that are generated from a band-limited function. 
Therefore a more general error analysis is required. 

To this end, we take the following strategy. First we show that since the network approximates smooth functions, then its spectrum decay fast. Then we use this to bound the error of the network for data that is generated from bandlimited mappings. 

\subsection{Spectral decay rate of networks with bounded weights}

We introduce the following lemma that provide a bound on the decay rate of finite neural networks with bounded weights. 

\begin{lem}
\label{lem:decay}
Let $\phi_n(x)$ be a finite multi-layer neural network with bounded weights (i.e., $\prod_{i=1}^L \norm{W_i}$ or $\prod_{i=1}^L \norm{W_i}_F$ are bounded), and non-expansive non-linearities that have a finite amount of discontinuities in their first derivative. Assume the training data is in the interval $[-\pi, \pi]$.
Then the Fourier coefficients of $\phi_n(x)$, 
obeys 
\begin{eqnarray}
\zeta_{k} = O(\abs{k}^{-2}).
\end{eqnarray}
\end{lem}

\emph{Proof.}
According to Corollary~\ref{cor:bounded_lap}, if the network has bounded weights then $\phi_n(x)$ has a finite
total variation of the derivative of this function, i.e., $\int_{x \in [-\pi, \pi]}  \phi_n''(x)dx$. Clearly, in this case also $\tilde{\phi}_n(x) = \phi(x) I_{[-\pi,\pi]}$ has a finite total variation. 
The indicator function $I_{[-\pi,\pi]}$ is one inside the domain $[-\pi,\pi]$ and zero outside of it. 

Notice that $\phi_n(x)I_{[-\pi,\pi]^d} \in L^1$. One way to see this is using the fact that $\phi_n(x)$ is Liphschitz (as it has a bounded first derivative as shown in Lemma~\ref{lem:bounded_norm}) and $[-\pi,\pi]$ is a finite domain. Thus, using a standard known result, the finite total variation in the first derivative implies that the Fourier transform $\hat{\phi}_n(w)$ of $\phi_n(x)I_{[-\pi,\pi]^d}$  obeys $\hat{\phi}_n(w) = O(\abs{w}^{-2})$.
Using the known relationship that the Fourier coefficients of $\phi_n$ ($\zeta_k$) are equal (up to a constant) to the ``sampled Fourier transform'' $\hat{\phi}_n(k)$ yields the desired result.
 \hfill $\Box$ 


Having the above decay rate for the Fourier coefficients of $\phi_n(x)$, we turn to bound the error between $\phi_n(x)$ and $f(x)$. 
We start with the case of uniform sampling and then move to the case of non-uniform sampling. 
We present both results for the univariate case. One may extend them to the multi-dimensional input case using a similar technique. We defer this to a future work.  

\subsection{Network error with uniform univariate samples}

The next theorem shows that the network error in the uniform sampling case decreases as a function of $\frac{1}{n^3}$.

\begin{thm}
\label{thm:uniform_sample_rec}
If a finite width univariate network has bounded weights (i.e., $\prod_{i=1}^L \norm{W_i}$ or $\prod_{i=1}^L \norm{W_i}_F$ are bounded), the training data of size $n \ge 2K+1$ is fitted by the network and it is uniformly sampled from a band-limited function with $2K+1$ non-zero Fourier coefficients, then we have 
\begin{eqnarray}
&& \norm{f(x) - \phi_n(x)}_{L^2_{[-\pi,\pi]}}^2 = \\ \nonumber && \int_{x=-\pi}^\pi (f(x) - \phi_n(x))^2 dx = O(1/n^3),
\end{eqnarray}
i.e., the error of the network scales as $O(1/n^3)$.
\end{thm}

The proof of this theorem is a special case of the one of Theorem~\ref{thm:nonuniform_sample_rec} for non-uniform sampling, which is presented next.

\subsection{Network error with random univariate samples}

Having the result for the uniform sampling case, we move to study the random sampling case. Analyzing this case is more important as it resembles in a closer way the case of real data, where we get labels for randomly sampled inputs. We show in this case the rate of convergence is of the order of $\frac{1}{\tilde{N}^3}$ ($\tilde{N} \le n$), where we assume that the random sampling pattern generates an operator $D\in \CC{n \times \tilde{N}}$ that is invertible with a condition number $\kappa$. Notice that this enables us to tradeoff the network error decay rate and the condition number of $D$. If $\tilde{N} = n$ we get the fastest decay rate but the condition number is very bad. Reducing $\tilde{N}$ improves the condition number but slows down the decay rate. We discuss the case, which is equivalent to $x_i \sim U[-\pi,\pi]$ (i.e., sampling from a uniform distribution in the domain $[-\pi, \pi]$), after the proof of the theorem. We claim that in that random sampling case, the network error scales as $\frac{1}{n^3}$, like in the deterministic uniform sampling case.

\begin{thm}
\label{thm:nonuniform_sample_rec}
If a finite width univariate network has bounded weights (i.e., $\prod_{i=1}^L \norm{W_i}$ or $\prod_{i=1}^L \norm{W_i}_F$ are bounded), the training data $(x_i, y_i)$ of size $n \ge 2K+1$ is randomly sampled from a band-limited function $f$ with $2K+1$ non-zero Fourier coefficients (i.e., $y_i = f(x_i)$), an operator $D\in \CC{n \times \tilde{N}}$ ($\tilde{N} \le n$) that corresponds to the sampling pattern that is invertible with a condition number $\kappa$, and the network $\phi_n$ fits the data, then with high probability 
\begin{eqnarray}
&& \norm{f(x) - \phi_n(x)}_{L^2_{[-\pi,\pi]}}^2 =  O(\kappa^2/\tilde{N}^3).
\end{eqnarray}
\end{thm}

\emph{Proof.}
Let  $\tilde{N} = 2\tilde{K}+1$ for $\tilde{K} \in \Integer$.\footnote{This assumption is used just for the simplicity of the presentation to perform a symmetric expansion of $c$. If $n$ is even we can just perform a non-symmetric expansion.} 
From the Parseval identity and the fact that $f$ is band-limited, we have 
\begin{eqnarray}
\label{eq:uniform_rec_error_parseval}
 \norm{f(x) - \phi_n(x)}_{L^2_{[-\pi,\pi]}}^2 &=& \sum_{k=-\infty}^{\infty}\abs{c_k - \zeta_k}^2 \\
 \nonumber  &\hspace{-1in} =& \hspace{-0.5in} \sum_{k \le \tilde{K}}\abs{c_k - \zeta_k}^2 + \sum_{\abs{k} > \tilde{K}}\abs{\zeta_k}^2.
\end{eqnarray}

To bound the network error, we need to bound the two terms in the rhs (right hand side) of the \eqref{eq:uniform_rec_error_parseval}. 

We start with the second term. Using Lemma~\ref{lem:decay}, we have that $\abs{\zeta_k} \le a /\abs{k}^2$ for some constant $a$. Thus, 
\begin{eqnarray}
\label{eq:d_k_sum_bound}
\sum_{\abs{k} > \tilde{K}}\abs{\zeta_k}^2 \le a \sum_{\abs{k} > \tilde{K}} \frac{1}{\abs{k}^4} &=& O\left(\frac{1}{\tilde{K}^3}\right) 
\\ \nonumber &=& 
 O\left(\frac{1}{\tilde{N}^3}\right),
\end{eqnarray}
where the first equality follows from the decay rate of the sum $\sum_{\abs{k} > \tilde{K}} \frac{1}{\abs{k}^4}$. Plugging \eqref{eq:d_k_sum_bound} in \eqref{eq:uniform_rec_error_parseval} leads to 
\begin{eqnarray}
\label{eq:nonuniform_rec_error_parseval}
 \norm{f(x) - \phi_n(x)}_{L^2_{[-\pi,\pi]}}^2 \le  
 \sum_{k \le \tilde{K}}\abs{c_k - \zeta_k}^2 + O\left(\frac{1}{\tilde{N}^3}\right).
\end{eqnarray}


Turning to bound the first term in the rhs of \eqref{eq:nonuniform_rec_error_parseval}, notice that from the assumption that the network fitted the training data, we have $f(x_i) =\phi_n(x_i)$ for $1 \le i\le n$. Using the Fourier series expansion of $\phi_n(x)$, we have that
\begin{eqnarray}
\label{eq:phi_x}
  \phi_n(x_i) = \sum_{k \in \Integer} \zeta_k e^{j k x_i} 
= \sum_{\abs{k} \le \tilde{K}}\zeta_k e^{j k x_i} + \sum_{\abs{k} > \tilde{K}}\zeta_k e^{j k x_i}.
\end{eqnarray}
Denote by $y$ the vector whose $i$th entry is $\phi_n(x_i)$, $D$ the operator that contains $\{e^{j k x_i}\}_{k=-K}^{K}$ in its rows, $\zeta$ the vector containing the coefficients $\zeta_k$, $k \le \tilde{K}$, and $y_{\backslash \tilde{K}} \in \CC{n}$ the vector whose $i$th entry is equal to $\sum_{\abs{k} > \tilde{K}}\zeta_k e^{j k x_i}$.
With this notation, we may write \eqref{eq:phi_x} in a vector form
\begin{eqnarray}
\label{eq:y_Dz_y_res}
y = D\zeta + y_{\backslash \tilde{K}},
\end{eqnarray}
Denote by $\zeta^l$ the vector that contains the set of coefficients 
$\zeta_{-\tilde{K} +l\tilde{N}}, \dots ,\zeta_{\tilde{K}+l\tilde{N}}$. Notice that each coefficient in $\zeta^l$ is multiplied in $y_{\backslash \tilde{K}}$ by the same complex exponent as in the multiplication between $D$ and $\zeta$ but with a factor $e^{jl\tilde{N}x_i}$. Thus, by denoting $L_l = \diag\left(e^{j\tilde{N}lx_1}, \dots, e^{j\tilde{N}lx_n}   \right)$, the diagonal matrix that contains these exponent factors, we may write $y_{\backslash \tilde{K}} = \sum_{l\ne 0} L_l D \zeta^l$.
Using the assumption that $D$ is invertible and $y=Dc$, we get from \eqref{eq:y_Dz_y_res} that
\begin{eqnarray}
\label{eq:c_zeta_sum_equality}
c = \zeta + \sum_{l\ne 0} D^\dag L_l D \zeta^l.
\end{eqnarray}
Notice that $||c-\zeta||_2^2 = \sum_{k \le \tilde{K}}\abs{c_k - \zeta_k}^2$, which is exactly the term we want to bound in \eqref{eq:nonuniform_rec_error_parseval}.
From \eqref{eq:c_zeta_sum_equality}, we have 
\begin{eqnarray}
\label{eq:DLDd_norm_bound}
&& \hspace{-0.3in} \norm{c - \zeta}_2^2 = \norm{\sum_{l\ne 0} D^\dag L_l D \zeta^l}_2^2 \\ && \nonumber \hspace{-0.2in} = \sum_{l \ne  0} ||D^\dag L_l D \zeta^l||_2^2 + \sum_{q \ne l,0}\sum_{l \ne  0} (D^\dag L_l D \zeta^l)^*D^\dag L_l D \zeta^q
\\ \nonumber && \hspace{-0.2in} \le \sum_{l \ne  0} ||D^\dag L_l D \zeta^l||_2^2   + \sum_{q \ne l,0}\sum_{l \ne  0} \norm{D^\dag L_l D \zeta^l}\norm{D^\dag L_l D \zeta^q}_2
\\ \nonumber && \hspace{-0.2in} \le \kappa^2 \sum_{l \ne  0} ||\zeta^l||_2^2  + \kappa^2 \sum_{q \ne l,0}\sum_{l \ne  0} \norm{\zeta^l}\norm{\zeta^q}_2,
\end{eqnarray}
where we use the Cauchy Schwartz inequality in the second step, and matrix norm inequalities in the last step, namely,
\begin{eqnarray}
\norm{D^\dag L_l D \zeta^l}_2 \le \norm{D^\dag}_2 \norm{L_l}_2 \norm{ D}_2 \norm{\zeta^l}_2 
\end{eqnarray}
with the fact that $\norm{D^\dag}_2 = 1/\sigma_{\min}(D)$, $\norm{D}_2 = \sigma_{\max}(D)$, $\norm{L_l}_2 = 1$ and $\kappa = \sigma_{\max}(D)/\sigma_{\min}(D)$.

We turn to bound the terms at the rhs of \eqref{eq:DLDd_norm_bound}. 
For the first, we have that  $\sum_{l \ne  0} ||\zeta^l||_2^2 = O\left( \frac{1}{\tilde{N}^3} \right)$ as in \eqref{eq:d_k_sum_bound}. For the second term, from Lemma~\ref{lem:decay}, we have that $\norm{\zeta^l}_2$ and $\norm{\zeta^q}_2$ behave as $\sqrt{\frac{n}{(nl)^4}}=\frac{1}{\sqrt{n}nl^2}$ and ${\frac{1}{\sqrt{n}nq^2}}$ respectively. Thus,
\begin{eqnarray}
\sum_{q \ne l,0}\sum_{l \ne  0} \norm{\zeta^l}_2\norm{\zeta^q}_2 \le \frac{1}{n^3} \sum_{q\ne 0} \frac{1}{q^2} \sum_{l \ne q} \frac{1}{l^2} 
= O\left(\frac{1}{n^3} \right),
\end{eqnarray}
where in the last equality we use the fact that $\sum_{l \ne q} \frac{1}{l^2} = constant$ and thus $ \sum_{q\ne 0} \frac{1}{q^2} \sum_{l \ne q} \frac{1}{l^2} =  constant $ as well.
Thus, we get from \eqref{eq:nonuniform_rec_error_parseval} that $\norm{c-\zeta}_2^2 = O(\kappa^2/\tilde{N}^3)$.  \eqref{eq:DLDd_norm_bound}. Combining this with \eqref{eq:nonuniform_rec_error_parseval} leads to the desired result.
 \hfill $\Box$ 

One may inquire what can be said on $\kappa$ in Theorem~\ref{thm:nonuniform_sample_rec}. To this end, we employ the empirical analysis performed in \cite{Haikin17Random}. In that work, it was conjectured that the eigenvalues of a randomly subsampled frame obey a Manova distribution. To employ their result in our case, we may treat $D$ as a matrix sampled from a significantly larger Fourier basis. In their work, they have two parameters. The first is $\gamma$, which is the fraction between the large basis and the size of the rows, namely $\tilde{N}$ in our case. This selection of subset of the rows creates a frame (the selection can be deterministic in this step of the selection as is our case). We set $\gamma = \epsilon \tilde{N}$, where $\epsilon$ is a very small number as the large basis should represent the whole space we are sampling from and we scale $\epsilon$ with $\tilde{N}$ as we get closer to the whole space when we add more samples. The second parameter is $\beta = \frac{n}{\tilde{N}}$, which is the redundancy factor in $D$. Given these two parameters, the support of the MANOVA distribution that characterize the singular values of $D$ is $[r_{-}, r_{+}]$, where
\begin{eqnarray}
r_{\pm} & = & \left( \sqrt{\beta(1-\gamma)} \pm \sqrt{1-\beta\gamma}   \right)^2
\\ \nonumber & = & \left( \sqrt{\frac{n}{\tilde{N}} - \epsilon n} \pm \sqrt{1-\epsilon n}   \right)^2.
\end{eqnarray}
Note that $r_{-}/r_{+}$ provides a bound to the condition number (as the minimal/maximal singular value may be greater/smaller than $r_{-}$/$r_{+}$). Assuming $\epsilon n$ is negligible, we have that
\begin{eqnarray}
\label{ref:kappa_bound}
\kappa \le \frac{(\sqrt{\beta} +1)^2}{(\sqrt{\beta}-1)^2}.
\end{eqnarray}
Notice that in Theorem~\ref{thm:nonuniform_sample_rec}, the ratio $\beta$ of $D$ is a free parameter that we may adjust to optimize the bound. This leads us to the following conjecture

\begin{conj}
If a finite width univariate network has bounded weights (i.e., $\prod_{i=1}^L \norm{W_i}$ or $\prod_{i=1}^L \norm{W_i}_F$ are bounded), the training data $(x_i, y_i)$ of size $n \ge 2K+1$ is randomly sampled from a band-limited function $f$ with $2K+1$ non-zero Fourier coefficients (i.e., $y_i = f(x_i)$), $x_i \sim U[-\pi,\pi]$, and the network $\phi_n$ fits the data, then 
\begin{eqnarray}
 \norm{f(x) - \phi_n(x)}_{L^2_{[-\pi,\pi]}}^2 =  O(1/n^3).
\end{eqnarray}
\end{conj}

It is a conjecture as it relies on empirical analysis  \cite{Haikin17Random} (with no rigorous proof) and on our assumptions above. If all of these are correct, then we get this result by simply plugging \eqref{ref:kappa_bound} and $\tilde{N} = \frac{n}{\beta}$ in the bound of Theorem~\ref{thm:nonuniform_sample_rec}, which yields
\begin{eqnarray}
 \norm{f(x) - \phi_n(x)}_{L^2_{[-\pi,\pi]}}^2 =  O\left(\frac{(\sqrt{\beta} +1)^4\beta^3}{(\sqrt{\beta}-1)^4}/n^3 \right).
\end{eqnarray}
Since $\beta$ is an arbitrary constant, the nominator can be also considered as such and thus we get that the error scales as $O(1/n^3)$. Notice that this bound is not tight and thus we cannot use it to approximate the ratio between the number of samples required in the deterministic and random cases in order to get the same error. Next we present a numerical simulation that demonstrates that this ratio is not so high and that both uniform deterministic and random sampling indeed obey a decay rate of $1/n^3$ for band-limited signals.

\begin{figure}[t]
\includegraphics[width=0.5\textwidth]{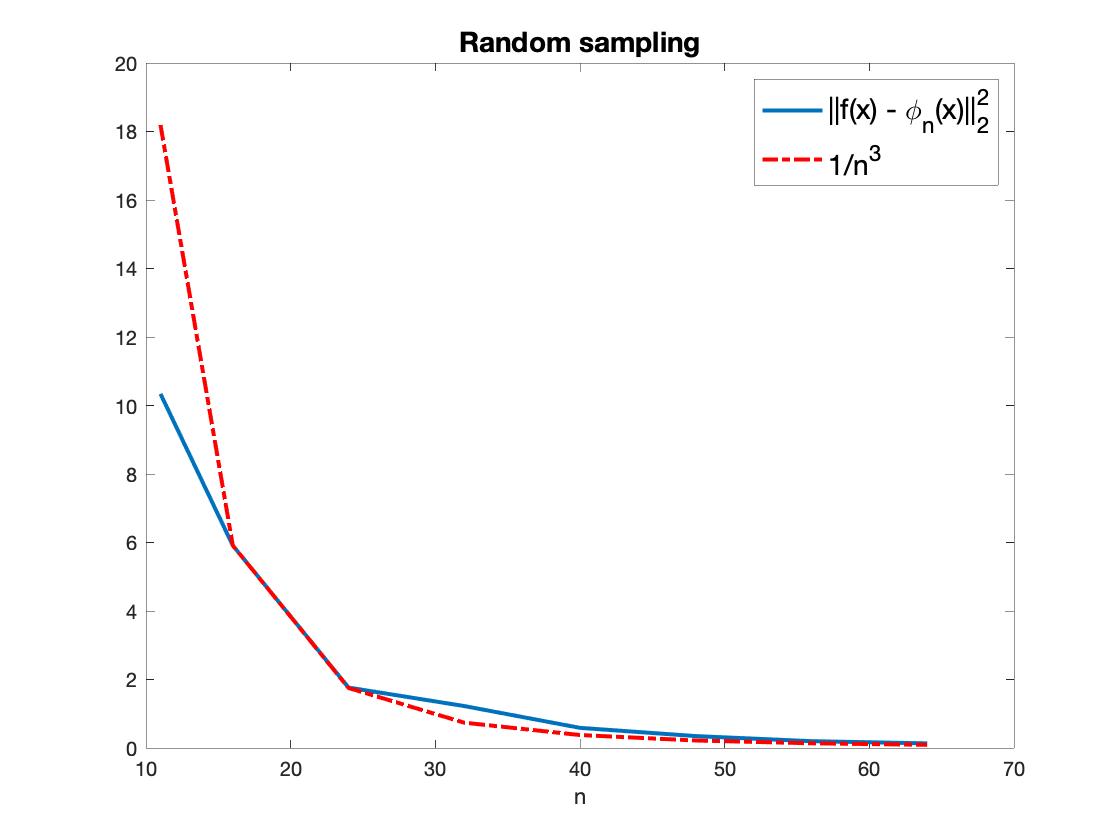}
\includegraphics[width=0.5\textwidth]{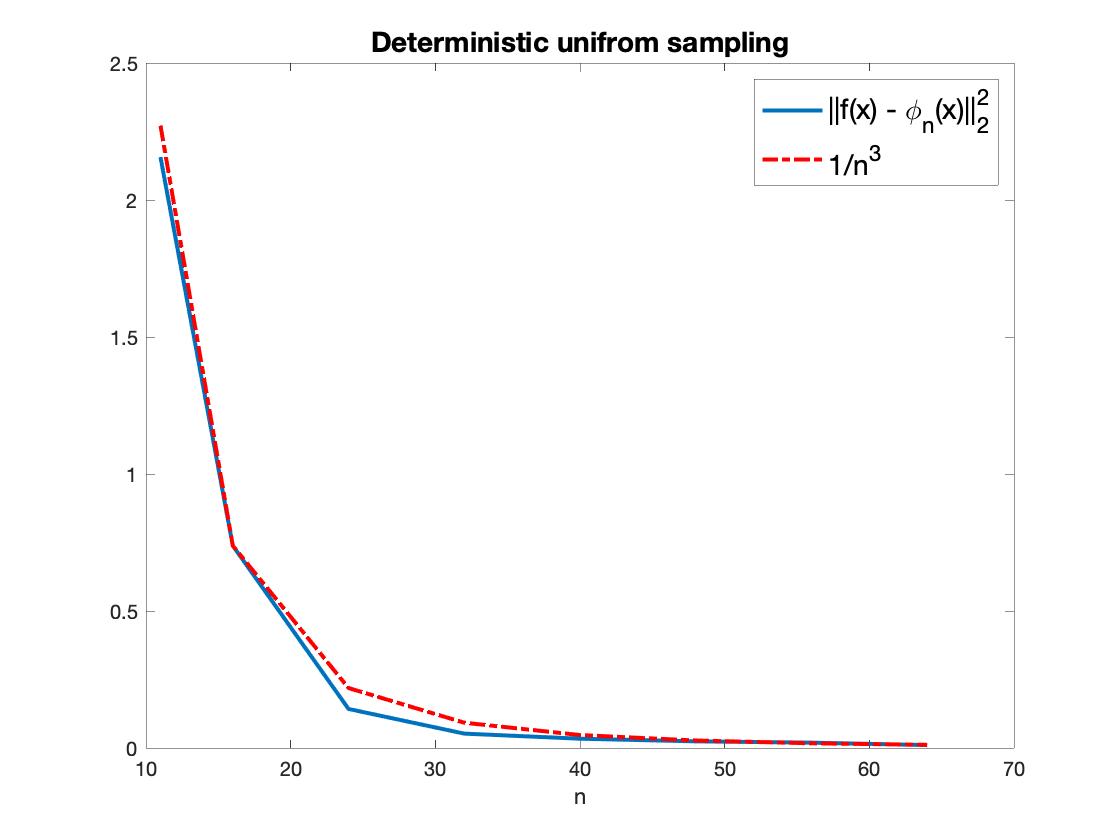}
\caption{\textit{Network error as a function of the number of training samples $n$.} Top: Training with random samples. Bottom: Training with uniform (equispaced) samples. We show in the small rectangles the same plots in log-log scale. Note that the network error scales as $1/n^3$ for both random and uniform cases.}
\label{fig:rec_error}
\end{figure}

\subsection{Network error with uniform multivariate samples}

\color{black}
We now turn to analyze the multivariate case. 
Yet, in this case, we limit ourselves to uniform sampling and to infinite width networks. We also use the following vector indexing notation. Given a vector of indices $k = [k_1, \dots, k_d]$, we abuse notation and use it to index functions and tensors by simply converting (uniquely) the tensor indices in $k$ to be vector indices as if the tensor/function was represented in a column-stack. 

We start with the following lemma that we use in our proof.
\begin{lem}
\label{lem:multi_decay}
If a two-layer neural network with a ReLU activation has an infinite width and bounded weights (assuming $\sum_{i=1}^2 \norm{W_i}_F^2$ is controlled) then the Fourier coefficients of $\phi_n(x)$ 
obeys 
\begin{eqnarray}
d_{ k} = O(\norm{k}_p^{-(d+1)})
\end{eqnarray}
for any $p \ge 1$.
\end{lem}

\emph{Proof.}
According to \cite{Ongie2020Function} (see Equations (10) and (11) there), if the network has bounded weights then the learned network $\tilde{\phi}_n(x)$ has a finite $\mathcal{R}$-norm. Clearly, in this case also $\tilde{\phi}_n(x)I_{[-\pi,\pi]^d} = \phi_n(x) I_{[-\pi,\pi]^d}$ has a finite $\mathcal{R}$-norm. 
The indicator function $I_{[-\pi,\pi]^d}$ is one inside the domain $[-\pi,\pi]^d$ and zero outside of it. 

Notice that $\phi_n(x)I_{[-\pi,\pi]^d} \in L^1$. One way to see this is using the fact that $\phi_n(x)$ is Liphschitz (e.g., see Proposition 8 in \cite{Ongie2020Function}) and $[-\pi,\pi]^d$ is a finite domain. Thus, from 
Proposition 12 in \cite{Ongie2020Function}, we have that the Fourier transform $\hat{\phi}_n(w)$ of $\phi_n(x)I_{[-\pi,\pi]^d}$  obeys $\hat{\phi}_n(tw) = O(\abs{t}^{-(d+1)})$.
Using the known relationship that the Fourier coefficients of $\phi_n$ ($d_k$) are equal (up to a constant) to the ``sampled Fourier transform'' $\hat{\phi}_n(k)$ and the fact that $\norm{t k}_p = t\norm{k}_p $ for any $p\ge 1$ yields the desired result.
\hfill $\Box$ 

Having the above decay rate for the Fourier coefficients of $\phi_n(x)$, we turn to bound the error between $\phi_n(x)$ and $f(x)$ in the multivariate case.

\begin{thm}
\label{thm:uniform_rhombus_rec}
If a two-layer multivariate neural-network with a ReLU activation has bounded weights as in Lemma~\ref{lem:multi_decay} and the training data is uniformly sampled on the $d$-dimensional grid such that $\norm{k}_1 \le K$, then we have 
\begin{eqnarray}
&& \norm{f(x) - \phi_n(x)}_{L^2_{[-\pi,\pi]^d}}^2 = \\ \nonumber && \int_{x \in [-\pi, \pi]^d} (f(x) - \phi_n(x))^2 dx = O(1/n^{(d+2)/d}),
\end{eqnarray}
i.e., the error of the network scales as $O(1/n^3)$
\end{thm}
\emph{Proof.}
Since we sample all points such that $\norm{k}_\infty \le K$, we have that $n = (2K+1)^d$.
From the Parseval identity and the fact that $f$ is band-limited, we have 
\begin{eqnarray}
\label{eq:uniform_rec_error_parseval_multivariate}
 \norm{f(x) - \phi_n(x)}_{L^2_{[-\pi,\pi]^d}}^2 &=& \sum_{k \in \Integer^d }\abs{c_k - d_k}^2 \\
 \nonumber  &\hspace{-1in} =& \hspace{-0.5in} \sum_{\norm{k}_\infty \le K}\abs{c_k - \zeta_k}^2 + \sum_{\norm{k}_\infty > K}\abs{d_k}^2.
\end{eqnarray}

To bound the network error, we need to bound the two terms in the rhs (right hand side) of the Eq.~\eqref{eq:uniform_rec_error_parseval_multivariate}. 

We start with the second term.  We have that
\begin{eqnarray}
\label{eq:d_k_sum_bound_multivariate}
&& \hspace{-0.25in}  \sum_{\norm{k}_{\infty} > K}\abs{d_k}^2 \le \sum_{\norm{k}_1 > K}\abs{d_k}^2 \le  a \sum_{\norm{k}_1 > K} \frac{1}{\norm{k}_1^{2(d+1)}} \\ \nonumber &&=
a \sum_{t \ge K+1} \sum_{\norm{k}_1 = t}  t^{-2(d+1)} 
\\ \nonumber && \le
a \sum_{t \ge K+1} 2^d{ t+d-1 \choose t} \frac{1}{t^{2(d+1)}} 
\\ \nonumber &&\le \frac{a2^d}{(d-1)!} \sum_{t \ge K+1} \frac{1}{t^{d+3}} \le O\left(\frac{2^d}{(d-1)!(K+1)^{d+2}}\right) \\ \nonumber && = O\left(\frac{4^d}{(d-1)!(2K+2)^{d+2}}\right) \le O\left(\frac{4^d}{(d-1)!n^{(d+2)/d}}\right),
\end{eqnarray}
where the first inequality is due to the fact that $\norm{k}_1 \ge \norm{k}_{\infty}$, the second inequality uses Lemma~\ref{lem:multi_decay} from which we have $\norm{d_k}_1 \le a /\norm{k}_1^2$ for some constant $a$, the following  equality use a simple split of the sum, the third inequality is due to a simple combinatorics identity for the sum of non-negative integers factored by $2^d$ to take into account each all orthants, 
the fourth inequality uses the fact that $(t+d-1)(t+d-2)\cdots (t+1)/t^{d-1} <1$ for large $t$, and the rest of the inequalities use standard arithmetic operations.

Turning to bound the first term in the rhs of Eq.~\eqref{eq:uniform_rec_error_parseval_multivariate}, notice that from the assumption that the network fitted the training data, we have $f(x_s) =\phi(x_s)$ for $s \in \Integer^d$ such that $\norm{s}_\infty \le K$. Using the Fourier series expansion of $\phi(x)$, we have that
\begin{eqnarray}
\label{eq:phi_x_multivariate}
&&  \phi_n(x_s) = \sum_{k \in \Integer^d} \zeta_k e^{j k^T x_s} \\ && \nonumber = \sum_{\norm{k}_\infty \le K}\zeta_k e^{j k^T x_s} + \sum_{\norm{k}_\infty > K}\zeta_k e^{j k^T x_s}.
\end{eqnarray}
Denote by $y$ the output that is equal at index $s$ to $\phi_n(x_s)$, $F\in \CC{n \times n}$ the $d$-dimensional DFT, $\zeta$ the vector containing the coefficients $\zeta_k$, $\norm{k}_\infty \le K$, and $y_{\backslash K} \in \CC{n}$ the vector whose $s$th entry is equal to $\sum_{\norm{k}_\infty > K}\zeta_k e^{j k^T x_s}$.
With this notation, we may write Eq.~\eqref{eq:phi_x_multivariate} in a vector form as we have done in Eq.~\eqref{eq:y_Dz_y_res}
\begin{eqnarray}
\label{eq:y_Fd_ybackslash_vector_multivariate}
y = F^* \zeta + y_{\backslash K}.
\end{eqnarray}
Using the fact that $\frac{1}{n}FF^* = I$ and $c=\frac{1}{n}Fy$ (as the samples are equispaced), we have 
\begin{equation}
    c = \zeta + \frac{1}{n}Fy_{\backslash {K}}.
\end{equation}
Moving $\zeta$ to the left hand side (lhs) and then taking a vector $\ell_2$ norm on both sides leads us to
\begin{equation}
\label{eq:c_d_y_backslashe_norm_multivariate}
    \norm{c - \zeta}_2^2=  \norm{\frac{1}{n}Fy_{\backslash {K}}}_2^2.
\end{equation}
Notice that $||c-\zeta||_2^2 = \sum_{\norm{k}_\infty \le K}\abs{c_k - \zeta_k}^2$, which is exactly the term we want to bound in \eqref{eq:uniform_rec_error_parseval_multivariate}. Thus, we just need to bound the rhs in \eqref{eq:c_d_y_backslashe_norm_multivariate}.

Because $x_s$ are uniform samples, we have $e^{j k^T x_s} = e^{j (k+(2K+1)q)^T x_s}$ for any $q\in \Integer^d$. Thus, we can partition the coefficients $\zeta_k$, $\abs{k}_\infty > K$ into groups of size $n = (2K+1)^d$. We can write each group as a vector $d^q$ whose entries contain the coefficients of $e^{j (k+(2K+1)q)^T x_s}$ for all $k$ such that $\norm{k}_\infty \le K$. With this notation, we have $y_{\backslash \tilde{K}} = \sum_{q \ne 0} F^* d^q $ (notice that we exclude $q=0$ as $d\zeta^0 = d$). Plugging it in Eq.~\eqref{eq:c_d_y_backslashe_norm_multivariate} leads to
\begin{equation}
    \norm{c - \zeta}_2^2=  ||\sum_{q \ne 0} d^q||_2^2.
\end{equation}
Expanding the rhs leads to 
\begin{eqnarray}
\label{eq:norm_sum_dl}
&& ||\sum_{q \ne 0} \zeta^q||_2^2 = \sum_{q \ne  0} ||\zeta^q||_2^2 + \sum_{l \ne q,0}\sum_{q \ne  0} (d^l)^*d^q
\\ \nonumber && \le  O\left(\frac{4^d}{(d-1)!n^{(d+2)/d}}\right) + \sum_{q \ne l,0} \norm{d^q}_2 \sum_{l \ne  0} \norm{d^l}_2,
\end{eqnarray}
where the bound for first term follows Eq.~\eqref{eq:d_k_sum_bound_multivariate} and for the second term we use Cauchy Schwartz inequality. Now notice that using Lemma~\ref{lem:multi_decay}, we have that $\norm{\zeta^l}_2$ and $\norm{\zeta^q}_2$ behave as $\sqrt{n\frac{1}{((2K+1)\norm{q}_1)^{2(d+1)}}}=\frac{1}{\sqrt{n^{(d+2)/d}}\norm{q}_1^{d+1}}$ and $\frac{1}{\sqrt{n^{(d+2)/d}}\norm{l}_1^{d+1}}$ respectively. Thus,
\begin{eqnarray}
&& \sum_{q \ne l,0}\sum_{l \ne  0} \norm{\zeta^l}_2\norm{\zeta^q}_2 \\ \nonumber && \le \frac{1}{n^{(d+2)/d}} \sum_{q\ne l,0} \frac{1}{\norm{q}_1^{d+1}} \sum_{l \ne 0} \frac{1}{\norm{l}_1^{d+1}} 
\\ \nonumber 
&&= O\left(\frac{1}{n^{(d+2)/d}} \right),
\end{eqnarray}
where in the last equality we use the fact that $\sum_{l \ne 0} \frac{1}{\norm{l}_1^{d+1}} = constant$  and thus $ \sum_{q\ne l,0} \frac{1}{\norm{q}_1^{d+1}} \sum_{l \ne 0} \frac{1}{\norm{l}_1^{d+1}} =  constant $ as well.
Thus, we get that $\norm{c-\zeta}_2^2 = O\left(\frac{1}{n^{(d+2)/d}} \right)$. Using this with Eq.~\eqref{eq:uniform_rec_error_parseval_multivariate} and the fact that $(d-1)!$ decays faster than $4^d$ leads to the desired result.

\hfill $\Box$

Extending this result to the random sampling case is possible in almost the same way done in the univariate case. Notice that this result for the multivariate case states that for large $d$ the squared error scales as $1/n$ (or the error without the square as $1/\sqrt{n}$), which coincide with classic generalization bounds that states that the error scales as $1/\sqrt{n}$. Yet, our result reveals the dependence on the dimension of the input. Thus, for small $d$ we get better error rates. 

\color{black}

\subsection{Beyond band-limited functions}

The underlying assumption in the above analysis is the mapping $f(x)$ is band-limited. Yet, one may inquire whether the input of neural networks really obeys this assumption. We claim here that we do not need to have this assumption in order to have the above generalization guarantees. 

Assume that the function $f(x)$ corresponds instead to the output of a neural network with bounded weights. As we have proven in Section~\ref{sec:net_func_space}, the spectrum of this function decays rapidly. This implies that most of the energy of this function is concentrated in a limited band and thus, we can repeat a similar derivation to the one performed above for such functions as well. Therefore, our assumption of band-limited functions is not restricting the implication of the theorem. 

\subsection{Empirical demonstration}

We have generated a bandlimited signal with $11$ Fourier coefficients ($K=5$). The signal is presented in Figure~\ref{fig:reconstruction}. We sampled both uniformly (equispaced) and randomly the function $f$, generating $n$ pairs of $(x_i, y_i = f(x_i))$, where $n \in \{11, 16,24,32,40, 48, 56, 64\}$. Then we trained a neural network with two hidden layers of size $1000$. We trained the network with weight decay and a SGD with momentum (with parameter 0.5). Once the network converged, we calculated its error compared to the generating function. Figure~\ref{fig:rec_error} shows that in both cases the error scales as $1/n^3$. The very larger error at $n=11$ in the random case may be explained by the fact that in this case, we can just have $\beta =1$ and then the condition number is relatively large, which increases the error. 

Notice that in the random sampling case, we need roughly twice the number of points to get to the same error as in the equispaced sampling case. This shows the great advantage of the latter. This observation may serve as a motivation for the farthest point sampling technique used in active learning when searching for new examples to annotate.

\begin{figure}
\begin{center}
   \includegraphics[width=0.5\textwidth]{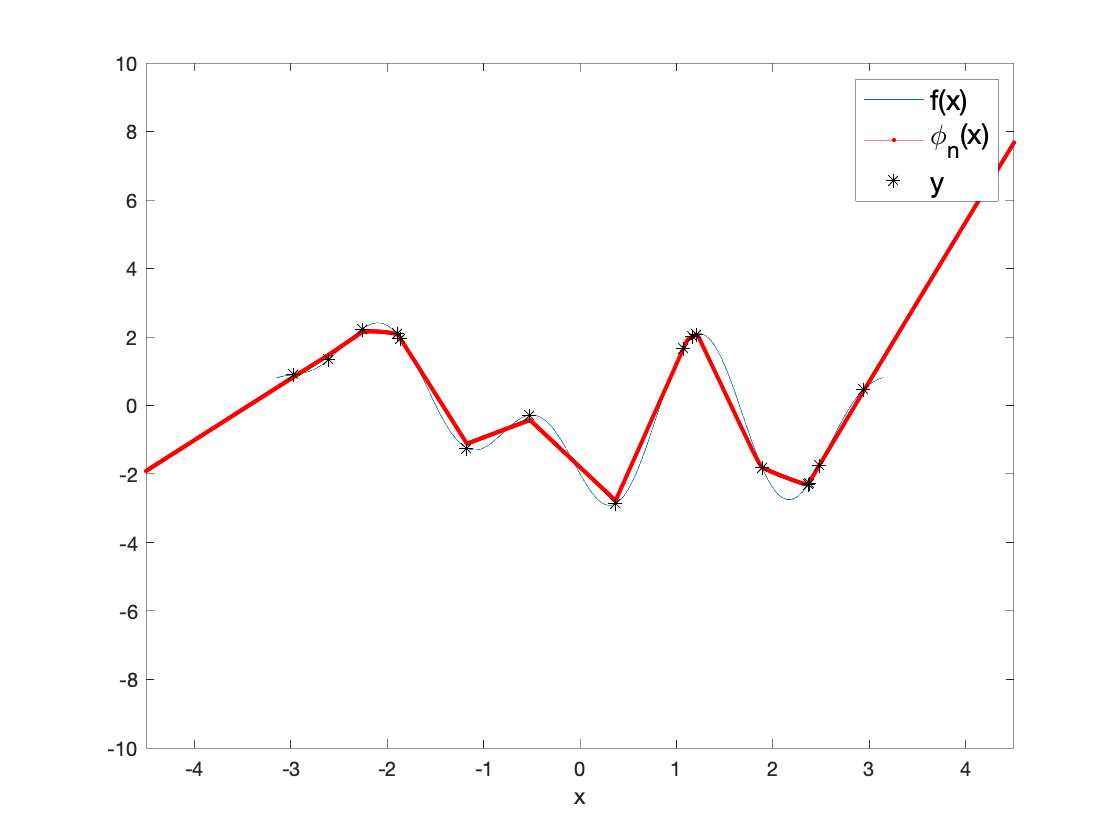}
\end{center}
   \caption{Approximation of a band-limited ($K=4$) function $f(x)$ using a network $\phi_n$ trained using only $16$ training examples ($y$).}
\label{fig:reconstruction_coef4}
\end{figure}

\begin{figure}
\includegraphics[width=0.5\textwidth]{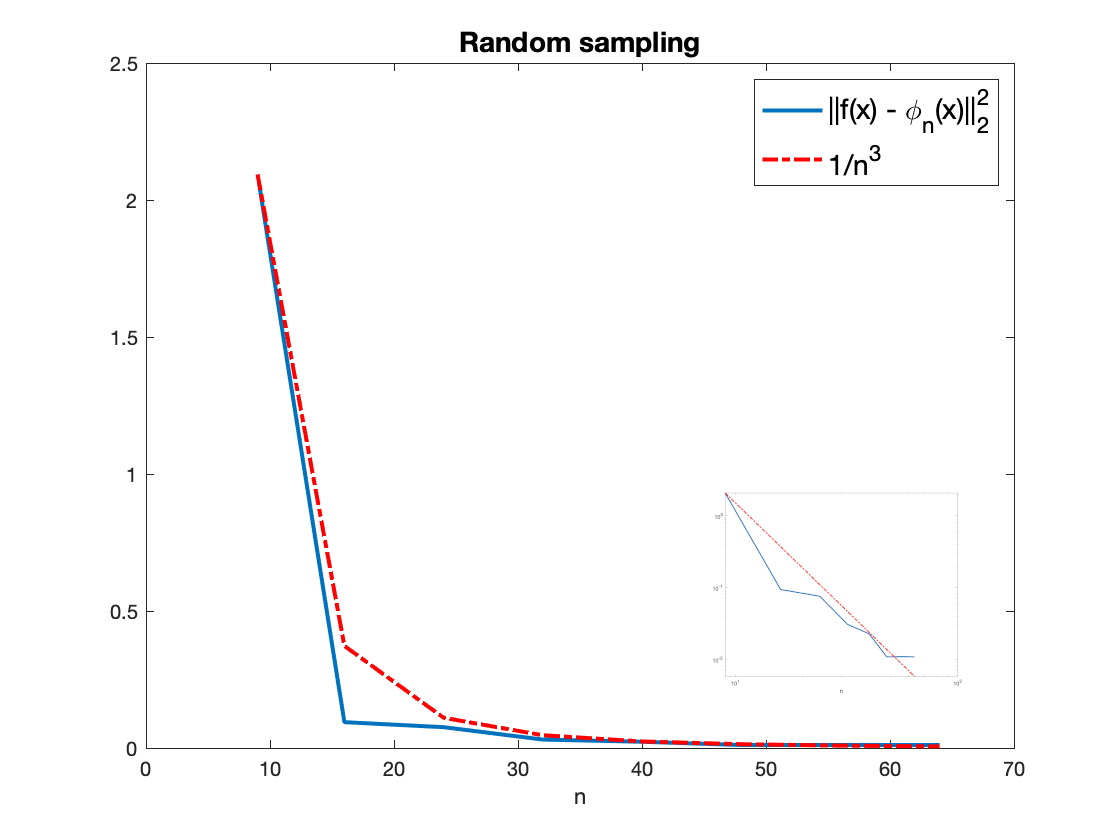}
\includegraphics[width=0.5\textwidth]{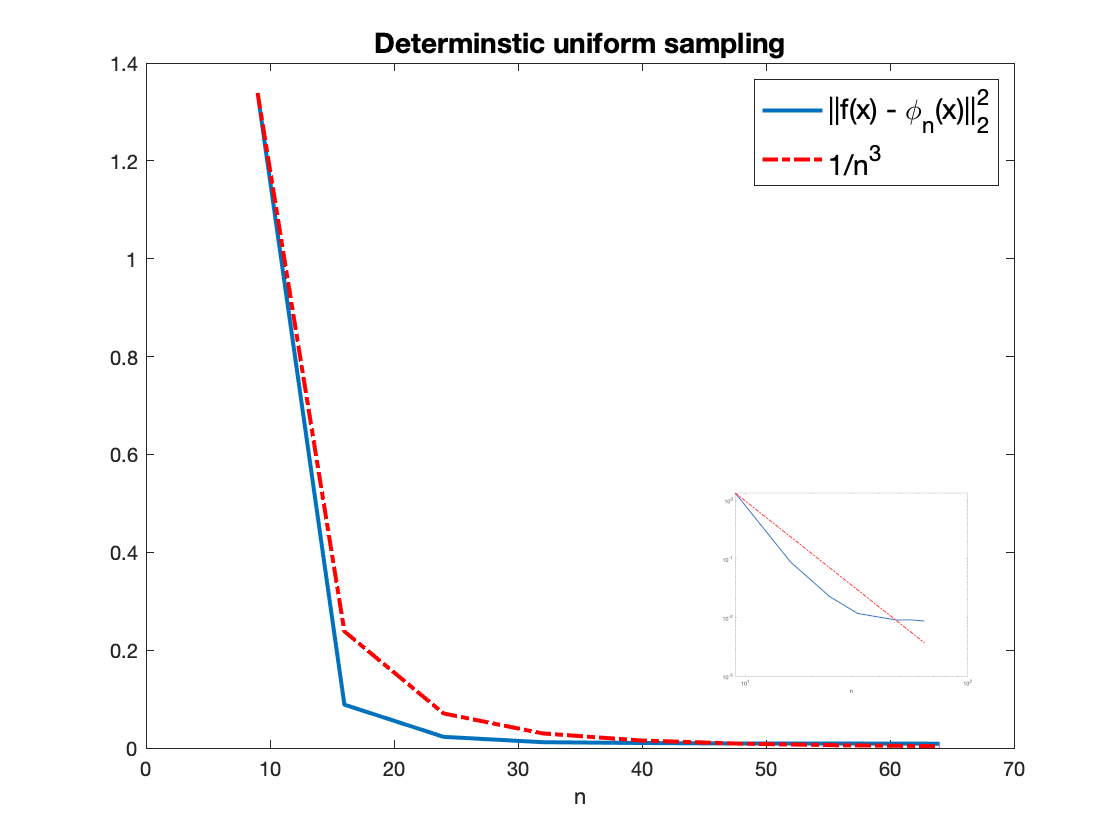}
\caption{\textit{Network error as a function of the number of training samples $n$ for $K=4$.} Top: Training with random samples. Bottom: Training with uniform (equispaced) samples. We put the same plots in a log-log scale in the small rectangles. Notice that the network error scales as $1/n^3$ in both cases.}
\label{fig:rec_error_coef4}
\end{figure}

In another experiment, we have generated a bandlimited signal with $9$ Fourier coefficients ($K=4$). The signal is presented in Figure~\ref{fig:reconstruction_coef4}. We sampled both uniformly (equispaced) and randomly the function $f$, generating $n$ pairs of $(x_i, y_i = f(x_i))$, where $n \in \{9, 16,24,32,40, 48, 56, 64\}$. Then we trained a neural network with two hidden layers of size $1000$. We trained the network with weight decay and a SGD with momentum (with parameter 0.5). Once the network converged, we calculated its error compared to the generating function. Figure~\ref{fig:rec_error_coef4} shows that in both cases the error scales as $1/n^3$ (the plateau at the end is probably due to numerical errors). As before, the larger error at $n=9$ in the random case may be explained by the fact that in this case, we can just have $\beta =1$ and then the condition number is relatively large, which increases the error. 

Notice that also here, in the case of a small number of samples, we get better error with deterministic uniform sampling compared to random sampling.

\section{Conclusion}
This work used sampling theory tools to analyze the error of neural networks. We showed that when the input data is band-limited, the network squared error scales as $O(1/n^{(d+2)/d})$. For the univariate case, we have shown that  the error scales as $1/n^3$ both with uniformly sampled and randomly sampled data. To the best of our knowledge, no such decay rate was demonstrated in the literature of neural network generalization (see for example the survey \cite{Jakubovitz2019Generalization}). As we assume that the network fits the data, the total network error studied in this work is the same as its generalization error.

While this work provides a generalization error of overparameterized networks with bounded weights, our analysis does not take into account the implicit bias on the margin of these networks implied by the optimization  
\cite{lyu2020gradient,Nacson19Lexicographic}, which is also important for network generalization \cite{Sokolic17Robust, Bartlett17Spectrally}. We believe that a combination of these tools may further improve the understanding of neural networks. 

In our work, we have assumed that the trained neural network is interpolating the data and that at this stage the weights are bounded. While this assumption holds in practice, in the theoretical works that prove interpolation of the training data by the network \cite{jacot2018neural,chizat2018note,mei2018mean,lee2019wide,arora2019exact} the $
\ell_2$ norm of the weights may increase with sample size for fitting non-smooth targets, possibly even exponentially in dimension for the Lipschitz case (see, e.g., \cite{Bach17Breaking}). Yet, as we focus on smooth target functions, the weights are likely to remain bounded as observed empirically. 
We defer to a future work to prove theoretically that these weights are bounded for networks that are trained with data generated from band-limited mapping, which we studied here. We believe that  for such band-limited target functions there will be a fixed upper bound depending on the bandwidth.

Notice that while the discussion in this paper was on band-limited functions, our results may be easily extended to other types of functions such as ones that have compact support in wavelets or splines. In this case, one may use tools from generalized sampling theory \cite{Jerri77Shannon, Unser00Sampling, Aldroubi01Nonuniform, Eldar2003Sampling, Eldar2015Sampling} to represent the signal in a similar way as we have done in \eqref{eq:y_Fc} and then perform a similar analysis to the one performed in this paper.

  \section*{Acknowledgments} The author would like to thank Tom Tirer and Dana Weitzner for fruitful discussion and the anonymous reviewers for their helpful remarks that significantly improved the paper. This work is supported by the  European research council starting grant (ERC-StG 757497 PI Giryes).

\bibliographystyle{IEEEtran}
\bibliography{sampling}

\begin{IEEEbiography}[{\includegraphics[width=1in,height=1.25in,clip,keepaspectratio]{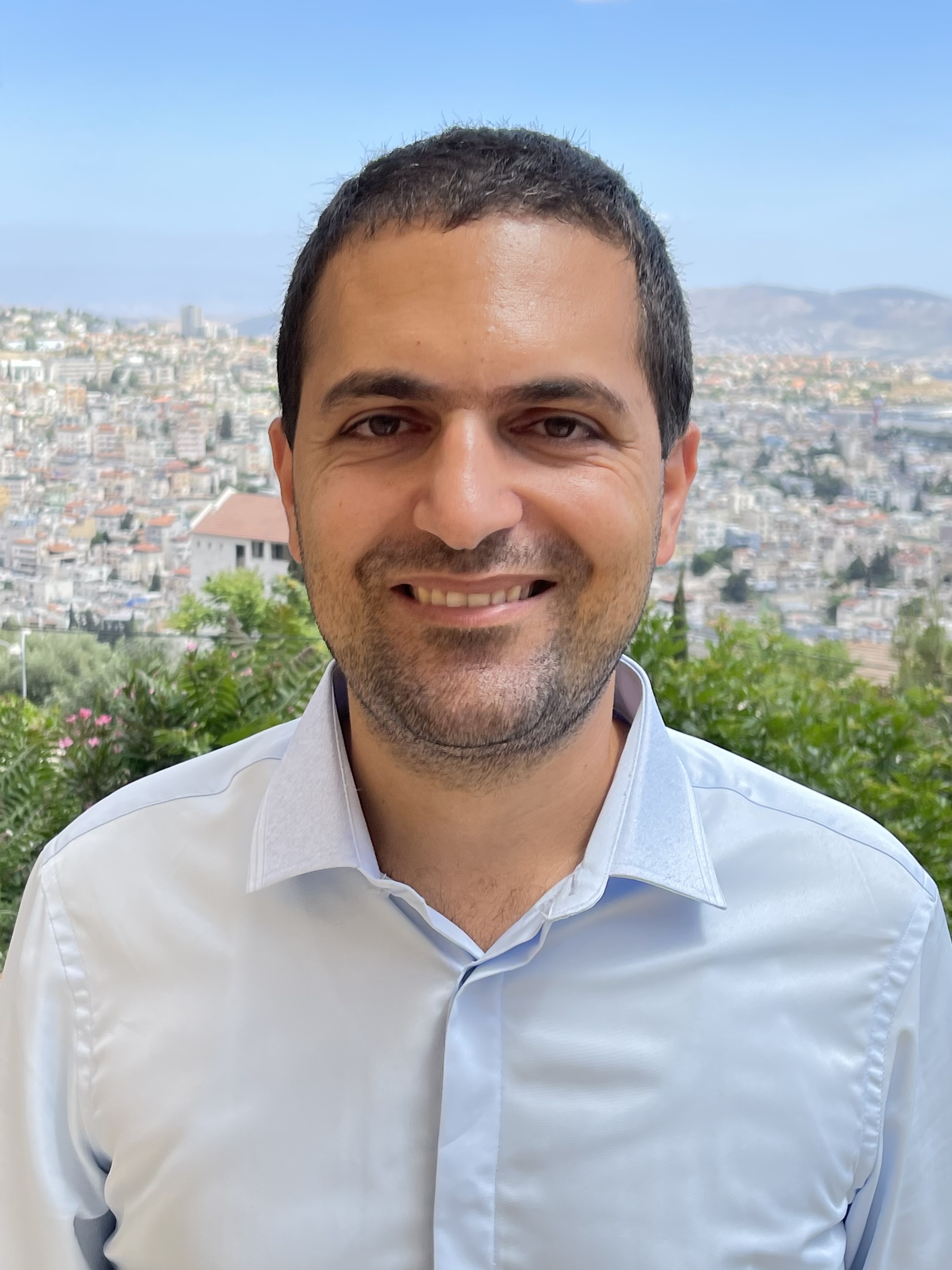}}]{Raja Giryes}
Raja Giryes is an associate professor in the school of electrical engineering at Tel Aviv University. His research interests lie at the intersection between signal and image processing and machine learning, and in particular, in deep learning, inverse problems, sparse representations, computational photography, and signal and image modeling. Raja received the EURASIP best P.hD. award, the ERC-StG grant, Maof prize for excellent young faculty (2016-2019), VATAT scholarship for excellent postdoctoral fellows (2014-2015), Intel Research and Excellence Award (2005, 2013), the Excellence in Signal Processing Award (ESPA) from Texas Instruments (2008) and was part of the Azrieli Fellows program (2010-2013). He is an associate editor in IEEE Transactions on Image Processing and Elsevier Pattern Recognition and has organized workshops and tutorials on deep learning theory in various conferences including ICML, CVPR, and ICCV. He serves as a consultant in various high-tech companies including Innoviz technologies and developed a technology that was used as the basis for the MultiVu technologies startup. 
\end{IEEEbiography}

\end{document}